\documentclass[journal]{IEEEtran}
\usepackage{graphicx}
\usepackage{amssymb}
\usepackage{amsfonts}
\usepackage{mathrsfs}
\usepackage{url}
\usepackage{amsmath}
\usepackage{algorithm}
\usepackage{algorithmic}
\usepackage{multirow}
\usepackage{booktabs}
\usepackage[table]{xcolor}
\usepackage{cite}
\usepackage{setspace}
\definecolor{mygray}{gray}{.9}
\definecolor{babyblueeyes}{rgb}{0.7, 0.8, 1}

\ifCLASSINFOpdf
\else
\fi

\begin{document}
	
	\title{Hyperspectral Image {{Classification}} with Markov Random Fields and a Convolutional Neural Network}
	
	\author{Xiangyong Cao$^{\dagger}$,~
		Feng Zhou,~\IEEEmembership{Member,~IEEE}, ~Lin Xu, ~Deyu Meng,~\IEEEmembership{Member,~IEEE}, ~Zongben Xu, \\and John Paisley~\IEEEmembership{Member,~IEEE}
		\thanks{Xiangyong Cao, Deyu Meng, and Zongben Xu are with the School of Mathematics and Statistics, Xi'an Jiaotong University, Xi'an 710049, China (caoxiangyong45@gmail.com, dymeng@mail.xjtu.edu.cn, zbxu@mail.xjtu.edu.cn).} 	
		\thanks{Feng Zhou is with the National Laboratory of Radar Signal Processing, Xidian University, Xi'an, China (fzhou@mail.xidian.edu.cn).} \thanks{Lin Xu is with the NYU Multimedia and Visual Computing Lab, New York University Abu Dhabi, UAE (xulinshadow@gmail.com).}
		\thanks{John Paisley is with the Department of Electrical Engineering \& Data Science Institute, Columbia University, New York, NY, USA (jpaisley@columbia.edu).}
		\thanks{This research was supported in part by National Basic Research Program (973 Program) of China under Grant No. 2013CB329404 the National Natural Science Foundation of China under Grants 91330204, 11131006 and 61373114.}
		\thanks{$\dagger$Xiangyong Cao performed this work at Columbia University on a grant from the China Scholarship Council.}
	}

	\maketitle
	
	\begin{abstract}
		This paper presents a new supervised {{classification}} algorithm for remotely sensed hyperspectral image (HSI) which integrates spectral and spatial information in a unified Bayesian framework. First, we formulate the HSI {{classification}} problem from a Bayesian perspective. Then, we adopt a convolutional neural network (CNN) to learn the posterior class distributions using a patch-wise training strategy to better use the spatial information. Next, spatial information is further considered by placing a spatial smoothness prior on the labels. {{Finally, we iteratively update the CNN parameters using stochastic gradient decent (SGD)} and update the class labels of all pixel vectors using $\alpha$-expansion min-cut-based algorithm.} Compared with other state-of-the-art methods, the proposed {{classification}} method achieves better performance on one synthetic dataset and two benchmark HSI datasets in a number of experimental settings.
	\end{abstract}
	
	\IEEEpeerreviewmaketitle
	\section{Introduction}
	Hyperspectral remote sensors capture digital images in hundreds of continuous narrow spectral bands to produce a high-dimensional hyperspectral image (HSI). Since HSI provides detailed information on spectral and spatial distributions of distinct materials{{~\cite{plaza2009recent}}}, it has been used for many applications, such as land-use mapping{{~\cite{camps2005kernel,petropoulos2012support,kitada2012land}}}, land-cover mapping{{~\cite{kitada2012land}}}, forest inventory{{~\cite{matsuki2015hyperspectral}}}, and urban-area monitoring{{~\cite{shafri2012hyperspectral}.}} All these applications require the material class label of each hyperspectral pixel vector and thus HSI classification has been an active research topic in the field of remote sensing. The aim of HSI classification is to categorize each hyperspectral pixel vector into a discrete set of meaningful classes according to the image contents.
	
	In the last few decades, many methods have been proposed for HSI classification. These methods can be roughly divided into two categories: spectral based methods and spectral-spatial based methods. We first briefly review these {{HSI classification approaches}} and then discuss the contributions of our proposed method.
	
	\subsection{Related work: spectral vs spectral-spatial based methods}
	Many classical HSI classification approaches are only based on spectral information~\cite{licciardi2012linear,villa2011hyperspectral,bandos2009classification}. {{Among}} these methods, {{pure spectral classification without any band reduction has often been proposed in the literature~\cite{civco1993artificial,bischof1998finding,melgani2004classification,camps2004robust}. Besides, some other pure spectral methods which extract spectral feature first using some feature extraction methods, such as principal component analysis~\cite{licciardi2012linear}, independent component analysis~\cite{villa2011hyperspectral} and linear discriminant analysis~\cite{bandos2009classification}, have also been proposed.}} {{However}}, these approaches only consider the spectral information and ignore the correlations among distinct pixels in the image, which tends to decrease their classification performance relative to those which consider both. In this paper we instead focus on designing a 
spectral-spatial based classification method.
	
	Spectral-spatial based methods can help improve the classification performance since they incorporate additional spatial information from the HSI. It has been {{often observed}} that spatial information is often as crucial as spectral information in the HSI classification task~\cite{plaza2009recent,fauvel2013advances,tarabalka2009spectral,qian2013hyperspectral}. Therefore, spectral-spatial methods have been proposed that additionally consider spatial correlation information. One approach is to extract the spatial dependence in advance using a spatial feature extraction method before learning a classifier. The patch-based feature extraction method is a representative example of this approach~\cite{chen2011hyperspectral,chen2013hyperspectral,soltani2015spatial,sun2014structured}. Here, features are extracted from groups of neighboring pixels using a subspace learning technique such as low-rank matrix factorization~\cite{xu2015spectral}, dictionary learning~\cite{chen2011hyperspectral,
soltani2015spatial} or subspace 
	clustering~\cite{jia2015spectral}. Compared with the original spectral vector, the features extracted by patch-based methods have higher spatial smoothness with some reduced noise~\cite{cao2017integration}. 
	
	Another popular spectral-spatial approach to incorporate spatial information is using a Markov random field (MRF) to post-process the classification map. MRF is an undirected graphical model~\cite{besag1974spatial} that has been applied in a variety of fields from physics to computer vision and machine learning. In particular, they have been widely used for image processing tasks such as image registration~\cite{zikic2010linear}, image restoration~\cite{bhatt1994robust}, image compression~\cite{reyes2007lossy} and image segmentation~\cite{chen2014semantic}. In the image segmentation task, MRFs encourage neighboring pixels to have the same class label~\cite{li2010semisupervised}. This has been shown to greatly improve the classification accuracy in HSI classification task~\cite{li2012spectral,tarabalka2014graph,cao2017integration}.
	
	{{Another family of spatial-spectral methods is based on spatial regularization (anisotropic smoothing) prior to spectral classification of the regularized image~\cite{mendez2012efficiency,duarte2008multiscale}.}} Aside from these methods, some other spatial-spectral methods have also been proposed, such as 3-dimensional discrete wavelet transform~\cite{cao2017integration,qian2013hyperspectral}, 3-dimensional Gabor wavelets~\cite{shen2011three}, morphological profiles~\cite{benediktsson2005classification}, attribute profiles~\cite{dalla2011classification} and {{manifold learning methods~\cite{crawford2009manifold,ma2010local}.}}
	
	{{Although the previous approaches perform well}}, a drawback to {{many of these approaches}} is that the extracted features are hand-crafted. Specifically, {{these approaches extract the features of the HSI by pre-specified strategies manually designed directly on data without using the label information and not via an ``end-to-end'' manner such as the deep learning approaches introduced here. Thus they}} highly depend on prior knowledge of the specific domain and are often sub-optimal~\cite{zhang2017spectral}. In the last few years, deep learning has been a powerful machine learning technique for learning data-dependent and hierarchical feature representations from raw data~\cite{Goodfellow-et-al-2016}. Such methods have been widely applied to image processing and computer vision problems, such as image classification~\cite{krizhevsky2012imagenet}, image segmentation~\cite{chen2014semantic}, action recognition~\cite{ji20133d} and object detection~\cite{szegedy2013deep}.
	
	Recently, a few deep learning methods have been introduced for the HSI classification task. For example, unsupervised feature learning methods such as stacked autoencoders~\cite{chen2014deep} and deep belief networks~\cite{chen2015spectral} have been proposed. Although these two unsupervised learning models can extract deep hierarchical feature, the 3-dimensional (3D) patch must be first flattened into 1-dimensional (1D) vectors in order to satisfy the input requirement, thus losing some spatial information. A supervised autoencoder~\cite{lin2013spectral} method that used label information during feature learning was also proposed. Similar methods have been proposed based on the convolutional neural network (CNN), for example, using a five-layer CNN trained on a 1D spectral vector without using spatial information~\cite{hu2015deep}, using a modified spectral-spatial deep CNN (SS-DCNN)~\cite{yue2015spectral} with 3D 
	patch input, or deep CNNs with spatial pyramid pooling (SPP)~\cite{he2014spatial} (SPP-DCNN)~\cite{yue2016deep}. Other methods based on CNNs~\cite{paisitkriangkrai2015effective,zhang2017spectral} and  recurrent neural networks (RNNs)~\cite{mou2017deep} have also been proposed.
 	
	\subsection{Contributions of our approach}
	As mentioned, compared with the traditional spectral-spatial HSI classification methods, deep learning can directly learn data-dependent and hierarchical feature representation from raw data. Although all the above deep learning methods obtain good performance, none of them formulate the HSI classification task into a Bayesian framework, where deep learning and MRF are considered simultaneously. Therefore, in this paper we propose a new supervised HSI {{classification}} algorithm in a Bayesian framework based on deep learning and MRF. Our contributions are threefold:
	
	1) We formulate the HSI {{classification}} problem from a Bayesian perspective and solve this problem by introducing an intermediate variable and other simplifications.
	
	2) {{We propose a new model which combines a CNN with a smooth MRF prior. Specifically, the CNN is used to extract spectral-spatial features from 3D patches and the smooth MRF prior is placed on the labels to further exploit spatial information. Optimizing the final model can be done by iteratively updating CNN parameters and class labels of all the pixel vectors. In this way, we integrate the MRF with the CNN. To our knowledge, this is the first approach that integrates the MRF with the CNN for the HSI classification problem.}}
	
	3) Our experimental results on one synthetic HSI dataset and two real HSI datasets demonstrate that the proposed method outperforms other state-of-the-art methods for the HSI {{classification}} problem, in particular the three deep learning methods SS-DCNN~\cite{yue2015spectral}, SPP-DCNN~\cite{yue2016deep} and DC-CNN~\cite{zhang2017spectral}.
	
	The rest of the paper is organized as follows: In Section 2 we formulate the HSI {{classification}} problem from a Bayesian perspective. In Section 3 we describe our proposed approach. In Section 4 we conduct experiments on one synthetic dataset and two benchmark real HSI datasets and compare with state-of-the-art methods. We conclude in Section 5. 
	
	\section{Problem Formulation}
	Before formulating the HSI {{classification}} problem, we first define the problem-related notations used throughout the paper.
	
	\subsubsection{Problem notation} Let the HSI dataset be $\mathbf{\mathcal{H}}\in R^{h\times w\times d}$, where $h$ and $w$ are the height and width of the spatial dimensions, respectively, and $d$ is the number of spectral bands. The set of class labels is defined as $\mathcal{K}=\{1,2,\dots,K\}$, where $K$ is the number of classes for the given HSI dataset. The set of all patches extracted from HSI data $\mathbf{\mathcal{H}}$ is denoted as $\mathbf{\mathcal{X}}=\{\mathbf{x}_{1},\mathbf{x}_{2},\dots,\mathbf{x}_{n}\}$, where $\mathbf{x}_{i}\in R^{k\times k\times d}$, $k$ is the patch size in the spatial dimension, $n=hw$ represents the total number of extracted patches (sliding step is 1 and padding is used). {{Each patch $\mathbf{x}_{i}$ is first fed into a CNN (shown in Fig.\ref{CNN}) and then a flattened vector $\mathbf{z}_{i}$ is output. The corresponding label of $\mathbf{z}_{i}$ is $y_i\in \mathcal{K}$, which is the label of the spectral vector corresponding to the center of $\mathbf{x}_{i}$.
 For the purpose of simplicity, we denote $(\mathbf{x}_i,y_i)$ as a sample, which doesn't mean all pixel vectors in $\mathbf{x}_i$ belong to the same class $y_{i}$ but means that the extracted spatial-spectral feature vector $\mathbf{z}_{i}$ belongs to class $y_i$. In other words, we use the 3-D patches only to obtain spatial-spectral feature for the corresponding central pixel vector of this patch.}} The label set is defined as $\mathbf{y}=\{y_1,y_2,\dots,y_n\}$. We define the training set for class $k$ as $\mathbf{\mathcal{D}}^{(k)}_{l^{(k)}}=\{(\mathbf{x}_1,y_1),\dots,(\mathbf{x}_{l^{(k)}},y_{l^{(k)}})\}$ and thus the entire training set can be denoted as $\mathbf{\mathcal{D}}_{l}=\{\mathbf{\mathcal{D}}^{(1)}_{l^{(1)}},\mathbf{\mathcal{D}}^{(2)}_{l^{(2)}},\dots,\mathbf{\mathcal{D}}^{(K)}_{l^{(K)}}\}$, where $l=\sum_{k=1}^{K}l^{(k)} \ll n$ is the total number of training patches.
	
	\subsubsection{Problem setup} The goal of HSI classification is to assign a label $y_i$ to {{the central pixel vector of}} each patch $\mathbf{x}_i, i = 1,2,\dots,n$. In the discriminative classification framework, the estimation of labels $\mathbf{y}$ for observations $\mathbf{\mathcal{X}}$ can be obtained by maximizing a distribution $\mathbb{P}(\mathbf{y}|\mathbf{\mathcal{X}},\boldsymbol{\Theta})$, {{where $\boldsymbol{\Theta}$ is the parameters of classifier.}} In our framework, we seek to combine the spatial modeling power of a Markov random field with the discriminative power of deep learning. Therefore, we let this distribution be of the form 
	\begin{eqnarray}\label{model1}
	\mathbb{P}(\mathbf{y}|\mathbf{\mathcal{X}},\boldsymbol{\Theta}) 
	&=& \sum_{\widetilde{\mathbf{y}}} \mathbb{P}(\mathbf{y},\widetilde{\mathbf{y}}|\mathbf{\mathcal{X}},\boldsymbol{\Theta}) \nonumber\\
	&=& \sum_{\widetilde{\mathbf{y}}} \mathbb{P}(\mathbf{y}|\widetilde{\mathbf{y}})\mathbb{P}(\widetilde{\mathbf{y}}|\mathbf{\mathcal{X}},\boldsymbol{\Theta}) \nonumber\\
	&=& \sum_{\widetilde{\mathbf{y}}} \mathbb{P}(\mathbf{y}|\widetilde{\mathbf{y}})\prod_{i=1}^n \mathbb{P}(\widetilde{\mathbf{y}}_i|\mathbf{x}_{i},\boldsymbol{\Theta}),
	\end{eqnarray}
	where $\widetilde{\mathbf{y}}=[\widetilde{\mathbf{y}}_{1}^T;\widetilde{\mathbf{y}}_{2}^T;\dots;\widetilde{\mathbf{y}}_{n}^T]\in R^{n\times K}$ is an intermediate variable, in which each $\widetilde{\mathbf{y}}_{i}\in R^{K}$ provides an initial probabilistic label {{pseudo-annotations}} for each patch $\mathbf{x}_{i}$. {{In our work, we define the distribution $\mathbb{P}(\widetilde{\mathbf{y}}_i|\mathbf{x}_{i},\boldsymbol{\Theta})$ as
	\begin{eqnarray}
	\mathbb{P}(\widetilde{\mathbf{y}}_i|\mathbf{x}_{i},\boldsymbol{\Theta})=
	\begin{cases}
		1, ~~ \widetilde{\mathbf{y}}_i=f(\mathbf{x}_{i};\boldsymbol{\Theta})\\
		0, ~~ otherwise
	\end{cases}
	\end{eqnarray}
	where $f$ is a classifier. After the pseudo-annotations $\widetilde{\mathbf{y}}$ is obtained.}} Then, a second classifier $ \mathbb{P}(\mathbf{y}|\widetilde{\mathbf{y}})$ takes this collection of {{pseudo-annotations $\widetilde{\mathbf{y}}$}} and outputs the {{classification label}} $\widehat{\mathbf{y}}=[\widehat{y}_{1};\widehat{y}_{2};\dots;\widehat{y}_{n}]$, which can be made as $\widehat{\mathbf{y}} = \arg\max_{\mathbf{y}\in\mathcal{K}^{n}} \mathbb{P}(\mathbf{y}|\mathbf{\mathcal{X}},\boldsymbol{\Theta})$.
	
	However, since a sum over all $\widetilde{\mathbf{y}}$ is computationally intensive, we simplify this in two ways for computational convenience. First, instead of only learning $\widehat{\mathbf{y}}$ {{by maximizing $\mathbb{P}(\mathbf{y}|\mathbf{\mathcal{X}},\boldsymbol{\Theta})$}}, we learn a point estimate of the pairs $(\widehat{\mathbf{y}},\widetilde{\mathbf{y}})$ {{by maximizing $\mathbb{P}(\mathbf{y},\widetilde{\mathbf{y}}|\mathbf{\mathcal{X}},\boldsymbol{\Theta})$}}. A goal is then to solve
	\begin{eqnarray}\label{MAP}
	(\widehat{\mathbf{y}},\widetilde{\mathbf{y}})=\underset{\mathbf{y},\widetilde{\mathbf{y}}}{\arg\max}\left\{\log\mathbb{P}(\mathbf{y}|\widetilde{\mathbf{y}}) + \sum_{i=1}^{n}\log\mathbb{P}(\widetilde{\mathbf{y}}_i|\mathbf{x}_{i},\boldsymbol{\Theta})\right\}.
	\end{eqnarray}
	Our second simplification takes the form of breaking the optimization problem in Eq.\ (\ref{MAP}) into two subproblems. First, we calculate label {{pseudo-annotations $\widetilde{\mathbf{y}}$ based on the classifier $f$. Then, given the annotation values $\widetilde{\mathbf{y}}$, the {{current classification results $\widehat{\mathbf{y}}$}} can be obtained by maximizing $\log\mathbb{P}(\mathbf{y}|\widetilde{\mathbf{y}})$. Thus Eq.\ (\ref{MAP}) can be solved by the following steps:
	\begin{eqnarray}\label{MAP1}
		\widetilde{\mathbf{y}}_i &=& f(\mathbf{x}_{i};\boldsymbol{\Theta}^{*}),~i=1,2,\dots,n
	\end{eqnarray}
	\begin{eqnarray}\label{MAP2}
		\widehat{\mathbf{y}} &=& \underset{\mathbf{y}\in\mathcal{K}^{n}}{\arg\max}~\log\mathbb{P}(\mathbf{y}|\widetilde{\mathbf{y}}).
	\end{eqnarray}
	Here it should be emphasized that classifier $f$ is first trained on the training data set $\mathbf{\mathcal{D}}_{l}$ and $\boldsymbol{\Theta}^{*}$ is the learned parameters of the classifier. After obtaining the current classification labels $\widehat{\mathbf{y}}$ on the whole HSI, next we train the classifier again using all the patches $\mathbf{\mathcal{X}}$ and their corresponding labels $\widehat{\mathbf{y}}$. The two steps are iteratively implemented until some criteria are satisfied. In this way, we can make full use of the unlabeled patches and thus more information can be incorporated in our framework.}}
	
	In this paper, in order to fully utilize both spatial and spectral information of HSI, we design a new approach based on this classification framework. Specifically, we firstly {{adopt a convolutional neural network (CNN) as the classifier $f$, which can not only help to extract spectral-spatial features for HSI, but also can provide label pseudo-annotations $\widetilde{\mathbf{y}}_i$ for each $\mathbf{x}_{i}$. Then, we place a smoothness prior on labels $\mathbf{y}$ to further enforce spatial consistency. In the following section, we will introduce these two aspects in detail.}} 
	
	\section{Proposed Approach}
	In this section, we first introduce the convolutional neural network (CNN) classifier for HSIs. Then, we further consider spatial information by placing a smoothness prior on the labels $\mathbf{y}$ and formulate the {{classification task}} as a labeling problem in an MRF. {{Finally, we summarize the proposed classification algorithm.}}
	
	\subsection{Discriminative CNN-based classifier}
	Convolutional neural networks (CNN) {{play an important role in many computer vision tasks.}} They consist of combinations of convolutional layers and fully connected layers. Compared with the standard fully connected feed-forward neural network (also called multi-layer perceptrons, MLP), CNNs exploit spatial correlation by enforcing a local connectivity pattern between neurons of adjacent layers to achieve better performance in many image processing problem, such as image denoising~\cite{xie2012image}, image super-resolution~\cite{dong2016image}, image deraining~\cite{fu2017clearing} and image classification~\cite{krizhevsky2012imagenet}.
	
\begin{figure}
		\centering
		\includegraphics[width=1.05\linewidth]{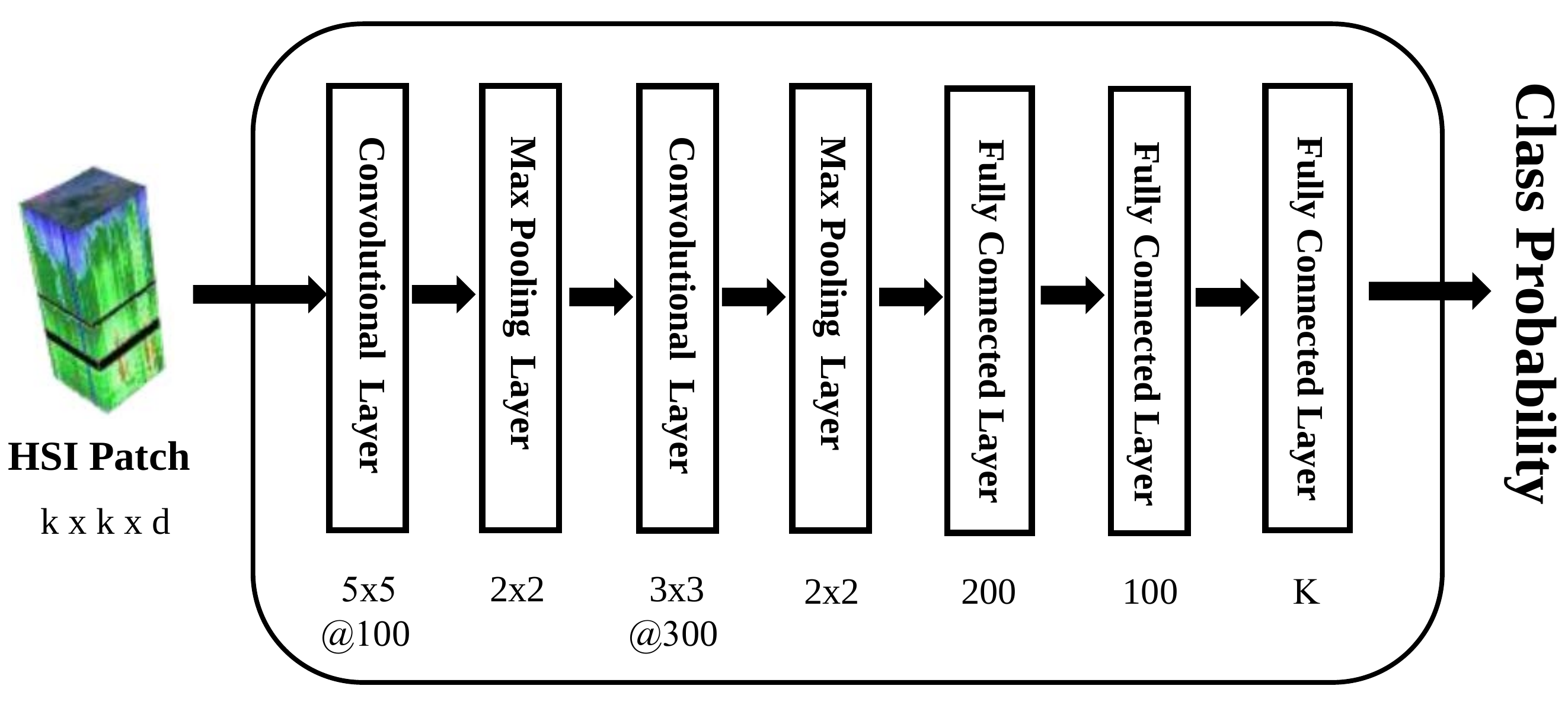}
		\caption{The network structure of the CNN used as our discriminative classifier.}\label{CNN}
	\end{figure}

	In this paper, we adopt the CNN structure shown in Figure~\ref{CNN}. This network contains one input layer, two pairs of convolution and max pooling layers, two fully connected layers and one output layer ({{The tuning of the network structure will be discussed in Section IV}}). The detailed parameter settings in each layer are also shown in Figure~\ref{CNN}.
	For the hyperspectral image classification task, each sample is a 3D patch of size $k\times k \times d$. Next, we introduce the flow of processing each sample patch $\mathbf{x}_{i}$ at each layer of the CNN.
	
	We note that there is no need to flatten the sample patch $\mathbf{x}_{i}$ into a 1-dimensional vector before it is fed into the input layer. Therefore, the input size of the first layer is $k\times k\times d$. First, the sample patch is input into the first convolutional layer, followed by a max pooling operation. The first convolutional layer filters the  $k\times k\times d$ sample patch using 100 {{filters}} of size {{$5\times 5\times d$}}. After this convolution, we have 100 feature maps each $n_{1}\times n_{1}$, where $n_{1}=k-4$. We then perform max pooling on the combined set of features of size $n_1\times n_1\times 100$. The kernel size of the max pooling layer is $2\times 2$ and thus the pooled feature maps have size of $n_2\times n_2\times 100$, where $n_2=\left\lceil n_{1}/2\right\rceil$. This set of pooled feature maps are then passed through a second pair of convolutional and max pooling layers. The convolutional layer contains 300 {{filters}} of size {{$3\times 3\times 100$}}, which filters 
the pooled feature maps into new feature maps of size $n_3\times n_3\times 300$, where $n_{3}=n_2-2$. Again these new feature maps are input into the second max pooling 
	layer with kernel size $2\times 2$ and turned into a second set of pooled feature maps of size $n_4\times n_4\times 300$, where $n_{4}=\left\lceil n_{3}/2\right\rceil$. Finally, the second pooled feature maps are flattened into a 1-dimensional vector $\mathbf{x}_{pool2}$ and input to the fully connected layer. The computation of the next three fully connected layers are as follows: 
	\begin{eqnarray}
	\mathbf{f}^{(5)}(\mathbf{x}_{pool2}) &=& \sigma(\mathbf{W}^{(5)}\mathbf{x}_{pool2}+\mathbf{b}^{(5)}),\\
	\mathbf{f}^{(6)}(\mathbf{x}_{pool2}) &=& \sigma(\mathbf{W}^{(6)}\mathbf{f}^{(5)}(\mathbf{x}_{pool2})+\mathbf{b}^{(6)}),\\
	\mathbf{f}^{(7)}(\mathbf{x}_{pool2}) &=&\mathbf{W}^{(7)}\mathbf{f}^{(6)}(\mathbf{x}_{pool2})+\mathbf{b}^{(7)},
	\end{eqnarray}
	where $\mathbf{W}^{(5)}$, $\mathbf{W}^{(6)}$ and $\mathbf{W}^{(7)}$ are weight matrices, $\mathbf{b}^{(5)}$, $\mathbf{b}^{(6)}$ and $\mathbf{b}^{(7)}$ are the biases of the nodes, and $\sigma(.)$ is the non-linear activation function. In our network, the activation functions in the convolutional layers and fully connected layers are all selected to be the rectified linear unit (ReLU) function. In order to simplify the notation, we denote $\mathbf{W}=\{\mathbf{W}^{(1)},\mathbf{W}^{(3)},\mathbf{W}^{(5)},\mathbf{W}^{(6)},\mathbf{W}^{(7)}\}$ and $\mathbf{b}=\{\mathbf{b}^{(1)},\mathbf{b}^{(3)},\mathbf{b}^{(5)},\mathbf{b}^{(6)},\mathbf{b}^{(7)}\}$, where $\{\mathbf{W}^{(1)},\mathbf{W}^{(3)}\}$ and $\{\mathbf{b}^{(1)},\mathbf{b}^{(3)}\}$ are the weight matrices and biases in the convolutional layers respectively. {{Thus the aforementioned classifier parameters $\boldsymbol{\Theta}$ are $\{\mathbf{W},\mathbf{b}\}$ in the CNN.}}
	
	The final vector $\mathbf{f}^{(7)}\in R^{K}$ is then passed to the softmax (normalized exponential) function, which gives a distribution on the label. As defined previously, the label {{pseudo-annotations}} is $\widetilde{\mathbf{y}}_{i}$ for a given sample $\mathbf{x}_{i}$. Then the {{pseudo-classification}} label can be obtained according to $y_{i}^{c}=\arg\max_{k\in\mathcal{K}} \widetilde{y}_{ik}$. Therefore, given the training set {{($\mathbf{\mathcal{D}}_{l}$ in the first updating and then the whole $(\mathbf{\mathcal{X}},\widehat{\mathbf{y}})$)}} where we represent each scalar label $y_i\in\mathcal{K}$ of a training sample as a one-hot vector $\mathbf{y}_{i}\in R^{K}$, the cross-entropy (CE) loss function can be computed as
	{{ \begin{eqnarray}\label{loss}
			E(\mathbf{W},\mathbf{b})&=& \frac{1}{l}\sum_{i}~CE(\mathbf{y}_{i},\widetilde{\mathbf{y}}_{i}^{(\mathbf{W},\mathbf{b})}),\nonumber \\
			&=& -\frac{1}{l}\sum_{i}\sum_{k=1}^{K}\mathbf{y}_{ik}\log{\widetilde{\mathbf{y}}_{ik}^{(\mathbf{W},\mathbf{b})}},~~~
			\end{eqnarray}}}where $\mathbf{W}$ and $\mathbf{b}$ is the parameter set defined above {{and here we denote $\widetilde{\mathbf{y}}_{i}$ as $\widetilde{\mathbf{y}}_{i}^{(\mathbf{W},\mathbf{b})}$ in order to emphasize that the pseudo-annotations $\widetilde{\mathbf{y}}_{i}$ is based on the learned CNN.}} The optimization of the loss function Eq.\ (\ref{loss}) is conducted by using the stochastic gradient descent (SGD) algorithm. In the $t_{th}$ iteration, the weight and bias are updated by
	\begin{eqnarray}
	\mathbf{W}_{t+1}&=& \mathbf{W}_{t} - \alpha\frac{\partial E(\mathbf{W},\mathbf{b})}{\partial \mathbf{W}}|_{\mathbf{W}_{t}},\\
	\mathbf{b}_{t+1}&=& \mathbf{b}_{t} - \alpha\frac{\partial E(\mathbf{W},\mathbf{b})}{\partial \mathbf{b}}|_{\mathbf{b}_{t}},
	\end{eqnarray}
	where the gradients with respect to $\mathbf{W}$ and $\mathbf{b}$ are calculated using the back-propagation algorithm~\cite{rumelhart1985learning} and $\alpha$ is the learning rate, which is set as 0.001 in our experiments. {{Here it should be noted that unlike some CNN applications, where the CNN parameters are first pre-trained on some samples prepared in advance and then fine-tuned on the new dataset, our proposed CNN-MRF method is directly applied to each dataset without pre-training. In all our experiments, the CNN weight parameters $\mathbf{W}$ are initialized by random standard normal distribution and the bias parameters $\mathbf{b}$ are initialized with zeros.}} In order to alleviate the issue of overfitting of the proposed CNN in the training phase, dropout strategy~\cite{srivastava2014dropout} is also adopted in the $5_{th}$ and $6_{th}$ fully connected layers and the dropout rate is set as 0.5 in our experiments.
	
	\subsection{Label smoothness prior and MRF optimization}
	After the label {{pseudo-annotations $\widetilde{\mathbf{y}}$ are obtained using the current trained CNN. Then, we begin to present our core model based on the simplifications from Eq. \ref{model1} to Eq. \ref{MAP2}.  Specifically, in order to obtain the classification results $\widehat{\mathbf{y}}$}} we need to solve the following optimization problem
	\begin{eqnarray}\label{segmodel}
	\widehat{\mathbf{y}} &=& \underset{\mathbf{y}\in\mathcal{K}^{n}}{\arg\max}~\log\mathbb{P}(\mathbf{y}|\widetilde{\mathbf{y}})\nonumber \\
	&=&\underset{\mathbf{y}\in\mathcal{K}^{n}}{\arg\max}~\log\mathbb{P}(\widetilde{\mathbf{y}}|\mathbf{y}) + \log\mathbb{P}(\mathbf{y})\nonumber \\
	&=&\underset{\mathbf{y}\in\mathcal{K}^{n}}{\arg\max}~\sum_{i=1}^{n}\log\mathbb{P}(\widetilde{\mathbf{y}}_{i}|{y}_{i}) + \log\mathbb{P}(\mathbf{y})
	\end{eqnarray}
	where $\widetilde{\mathbf{y}}$ can be regarded as new features for this problem, $\log\mathbb{P}(\widetilde{\mathbf{y}}_{i}|{y}_{i})$ is the log-likelihood and $\mathbb{P}(\mathbf{y})$ is the label prior. Specifically, in our work, the log-likelihood $\log\mathbb{P}(\widetilde{\mathbf{y}}_{i}|{y}_{i})$ is defined as
	\begin{eqnarray}\label{single_prior}
	\log\mathbb{P}(\widetilde{\mathbf{y}}_{i}|{y}_{i})=\sum_{k=1}^{K}1\{y_{i}=k\}
	\log{\widetilde{{y}}_{ik}},
	\end{eqnarray}
	where $1\{\cdot\}$ is an indicator function. 
	
	In image segmentation tasks, it is probable that adjacent pixels have the same label. The exploitation of this naive prior information can often dramatically improve the segmentation performance. In this paper, we enforce a spatially smooth prior on labels $\mathbf{y}$ to encourage neighboring pixels to belong to the same class. This smoothness prior distribution on labels $\mathbf{y}$ is defined as  
	\begin{eqnarray}\label{labelprior}
	\mathbb{P}(\mathbf{y}) = \frac{1}{Z}e^{\mu\sum_{i=1}^{n}\sum_{j\in\mathcal{N}(i)}\delta(y_{i}-y_{j})},
	\end{eqnarray} 
	where $Z$ is a normalization constant for the distribution, $\mu$ is the label smoothness parameter, $\mathcal{N}(i)$ is the neighboring pixels of pixel $i$ and $\delta(\cdot)$ is a function defined as: $\delta(0)=1$ and $\delta(y)=-1$ for $y\neq 0$. We note that the pairwise interaction terms $\delta(y_{i}-y_{j})$ obtain higher probability when neighboring labels are equal than when they are not equal. In this way, this smoothness prior can encourage piecewise smooth segmentations.
	
	Based on Eq.\ (\ref{single_prior}) and Eq.\ (\ref{labelprior}), the final {{classification}} model (\ref{segmodel}) is thus given by 
	\begin{equation}\label{final_seg}
	\widehat{\mathbf{y}}\!=\!\underset{\mathbf{y}\in\mathcal{K}^{n}}{\arg\max}\!\left\{\sum_{i=1}^{n}\!\sum_{k=1}^{K}\!1\{y_{i}=k\}\!
	\!\log{\widetilde{{y}}_{ik}}\!+\!\mu\!\sum_{i=1}^{n}\!\!\sum_{j\in\mathcal{N}(i)}\!\!\delta(y_{i}\!-\!y_{j})\right\}.
	\end{equation}
	This objective function contains many pairwise interaction terms and is a challenging combinatorial optimization problem. It can also be regarded as an MRF model in which an undirected model graph $\mathcal{G}=<\mathcal{V},\mathcal{E}>$ is defined on the whole image, where graph node sets $\mathcal{V}$ correspond to pixels, and the undirected edge set $\mathcal{E}$ represents the neighboring relationship between the pixels~\cite{boykov2001interactive}. We define a random variable $y_i$ on each node $v_{i}\in \mathcal{V}$, thus allowing the labels $\mathbf{y}$ to form a Markov random field. The objective function is its energy function, of which the first term represents the cost of a pixel being assigned with different classes. The larger the probability of a pixel belonging to a certain class, the more probable that the pixel is assigned the corresponding label. The second term of the energy function encourages the labels of neighboring pixels to be the same. 
	
	This optimization problem of the MRF is NP-hard. Many approximating algorithms have been proposed, such as the graph cut method~\cite{boykov2001fast,kolmogorov2004energy}, belief propagation~\cite{yedidia2005constructing} and message passing~\cite{kolmogorov2006convergent}. In this paper, we use the $\alpha$-expansion min-cut method~\cite{boykov2001fast} because of its good performance and fast computation speed. 
	
\begin{figure}[t]
			\centering
			\includegraphics[width=0.7\columnwidth]{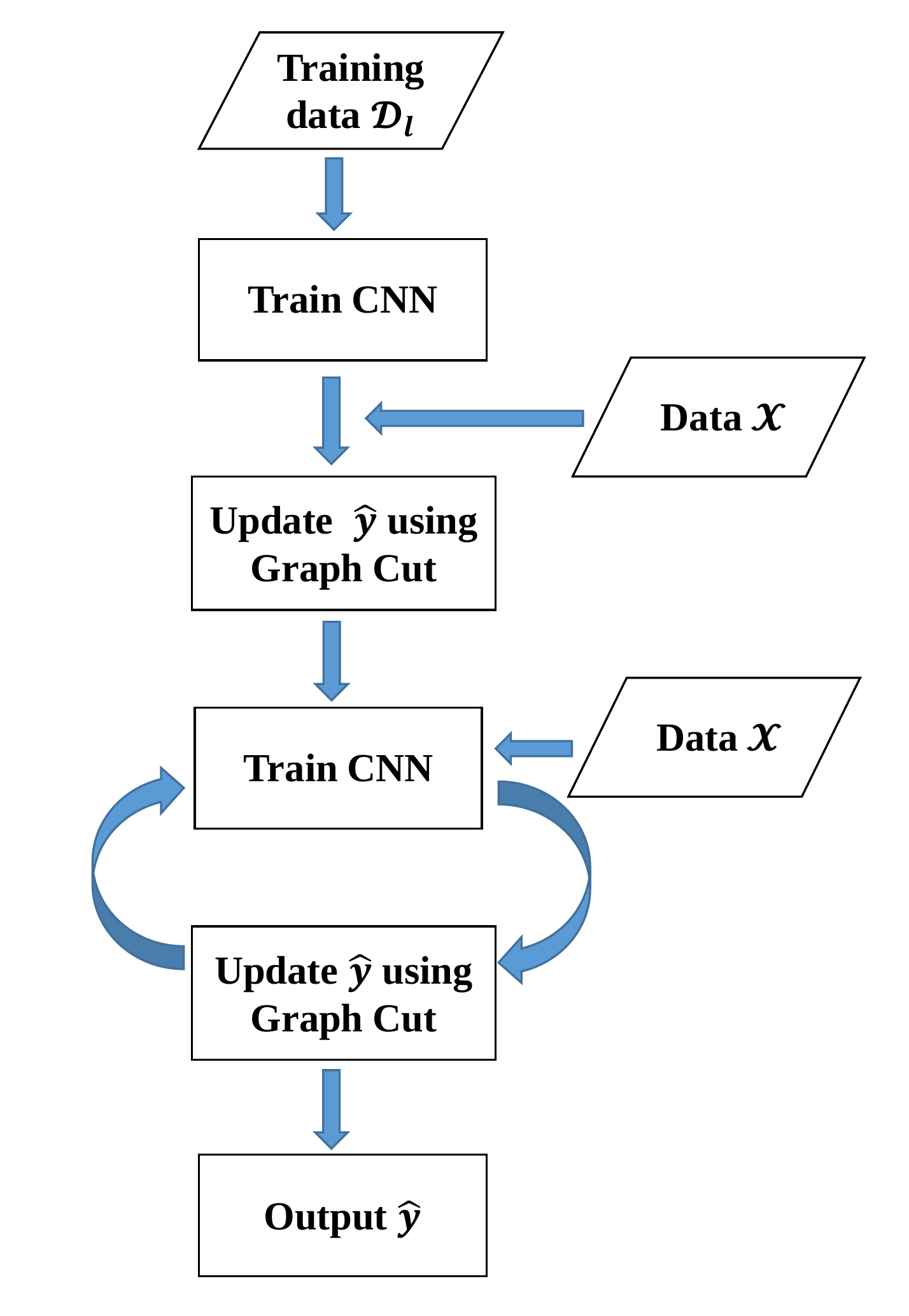}
			\caption{{{The flowchart of the proposed CNN-MRF algorithm.}}}\label{flowchart}
		\end{figure}

	{{
			\subsection{CNN-MRF classification algorithm}
			After introducing how to update CNN's parameters and compute class labels respectively, we can summarize our CNN-MRF classification algorithm in detail in Algorithm~\ref{alg2}. Besides, the corresponding flowchart of this algorithm is also depicted in Figure~\ref{flowchart}. 
		}}
		\begin{algorithm}[H]
			\caption{{{CNN-MRF classification algorithm for HSI}}} \label{alg2}
			\begin{algorithmic}[1]
				\REQUIRE HSI patches $\mathbf{\mathcal{X}}$, training data $\mathbf{\mathcal{D}}_{l}$, learning rate $\alpha$,\\ ~~~~ smoothness parameter $\mu$, batch size $n_{batch}$.
				\ENSURE Labels $\widehat{\mathbf{y}}$.
				{{
						\STATE Train the CNN classifier using $\mathbf{\mathcal{D}}_{l}$;\\
						\STATE Compute $\widetilde{\mathbf{y}}$ on $\mathcal{X}$ and update label $\widehat{\mathbf{y}}=\alpha$-Expansion($\widetilde{\mathbf{y}}$,$\mu$).\\
						\STATE Train the CNN classifier using $\{\mathcal{X},\widehat{\mathbf{y}}\}$; \\
						\STATE Compute $\widetilde{\mathbf{y}}$ on $\mathcal{X}$ and update label $\widehat{\mathbf{y}}=\alpha$-Expansion($\widetilde{\mathbf{y}}$,$\mu$).\\
						\STATE Repeat step 3 and 4 until the stopping criterion is satisfied.}} 
			\end{algorithmic}
		\end{algorithm}
		
		\section{Experiments}
		To test the effectiveness of our proposed CNN-MRF {{classification}} algorithm in different scenarios, we conduct experiments on one synthetic dataset and two real-world benchmark datasets.\footnote{Code is available at: \url{https://github.com/xiangyongcao/CNN_HSIC_MRF}.} For comparison, we consider several state-of-the-art HSI {{classification}} methods, including support vector machine graph-cut method (SVM-GC), subspace multinomial logistic regression with multilevel logistic prior (MLRsubMLL)\footnote{Code is available at: \url{https://www.lx.it.pt/~jun/demos.html}}~\cite{li2012spectral} and support vector machine based on the 3-dimensional discrete wavelet transform method ({{SVM-3DDWT-GC}})\footnote{Code is available at: \url{https://github.com/xiangyongcao/3DDWT-SVM-GC}}~\cite{cao2017integration}. {{It should be noted that the competing methods SVM-GC and MLRsubMLL are methods that integrate the MRF into the SVM and MLRsub methods, respectively. In order 
to distinguish the classification methods with MRF and without MRF, we denote them as classification methods and regularized classification methods, respectively.}} Besides, we also compare with three deep learning methods: SS-DCNN~\cite{yue2015spectral}, SPP-DCNN~\cite{yue2016deep} and DC-CNN~\cite{zhang2017spectral}. Our proposed CNN-MRF approach is implemented in Python using the Tensorflow~\cite{abadi2016tensorflow} library on a server with Nvidia 
		GeForce GTX 1080 and Tesla K40c. All non-deep algorithms are run in Matlab R2014b. For comparisons with other deep learning models, we use the best reported results for these algorithms. 
		
		All methods are compared numerically using the following three criteria~\cite{cao2017integration}: overall accuracy (OA), average accuracy (AA) and the kappa coefficient ($\kappa$). OA represents the number of correctly classified samples divided by the total number of test samples, AA denotes the average of individual class accuracies, and $\kappa$ involves both omission and commission errors and gives a good representation of the the overall performance of the classifier. For all the three criteria, a larger value indicates a better classification performance.
		
		\subsection{Synthetic HSI data}
		
		{{In this section, we first generate a synthetic HSI dataset. Then, we use this dataset to evaluate in detail the sensitivity of performance to different parameters setting of our CNN structure. Finally, we compare our method with other competing methods on this dataset using the tunning CNN structure.}}
		
		\subsubsection{Generation of Synthetic HSI data}
		To generate a synthetic HSI data, five endmembers are first extracted  randomly from a real scene with 162 bands in ranges 400$-$2500 nm, and then 40000 vectors are generated as a sum of Gaussian fields with constraints so as to respect the abundance-nonnegative-constraint (ANC) and abundance-sum-to-one-constraint (ASC). Finally, this dataset is generated using a Generalized Bilinear Mixing Model (GBM)\cite{zhu2017unsupervised}:
		\begin{eqnarray}
		\mathbf{z}&=& \sum_{i=1}^{K}a_{i}\mathbf{e}_{i}+\sum_{i=1}^{K-1}\sum_{j=i+1}^{K}\gamma_{ij}a_{i}a_{j}\mathbf{e}_{i}\odot\mathbf{e}_{j}+\mathbf{n},
		\end{eqnarray}
		with class number $K=5$. Here $\mathbf{z}$ is the simulated pixel vector, $\gamma_{ij}$ are selected uniformly from $[0,1]$, $\mathbf{e}_{i}, i=1,\dots,5$ are the five end-members, $\mathbf{n}$ is the Gaussian noise with an SNR of 30 dB, $a_{i}\geq 0$ and $\sum_{i=1}^{K}a_{i}=1$. 
		
		
		\subsubsection{Impact of parameter settings}
		In this section, we evaluate in detail the sensitivity of performance to different parameters settings {{of our CNN structure}}. In the following experiments, we use {{the synthetic dataset}} and randomly choose 1\% training samples from each class as training data and the remaining for testing.
		
		(1) \textit{Kernel Size}: First, we test the impact of different kernel sizes. We fix the kernel size of the second layer as 3 and evaluate the performance by varying the kernel size in the first layer. Table \ref{kernelsize} shows these results. From this table we can conclude that larger kernel sizes can obtain better results. This is because more structure and texture can be captured using a larger kernel size. Therefore, in our experiments, we set the kernel size of the first layer as 5.  
		
		\begin{table}[htp]
			\caption{\label{kernelsize} {{Overall accuracy (\%) with different kernel sizes.}}}
			\begin{center}
				{\normalsize
					\scalebox{0.85}[0.85]
					{
						\begin{tabular}{|c|c|c|c|c|c|}
							\hline
							Kernel Size & 1      & 2     & 3     & 4     &  5 \\
							\hline
							OA      & 96.07  & 96.71 & 97.10 & 98.26 & \bf{99.51} \\
							\hline
						\end{tabular}
					}
				}
			\end{center}
		\end{table}

		(2) \textit{Network Width}: Next, we evaluate the impact of network width in the convolutional layer on the {{classification}} results. Fixing the network width of the first convolutional layer to 100, we test the performance by changing the width of the second convolutional layer. These results are displayed in Table \ref{networkwidth}. It can be observed that the results are not very sensitive to the network width and thus in our experiments, we select 200 as the default setting of the network width in the second convolutional layer.
		
		\begin{table}[htp]
			\caption{\label{networkwidth} {{Overall accuracy (\%) with different network widths.}}}
			\begin{center}
				{\normalsize
					\scalebox{0.85}[0.85]
					{
						\begin{tabular}{|c|c|c|c|c|c|}
							\hline
							Network Width  & 50    & 100    & 200    & 300   &  500 \\
							\hline
							OA            & 99.27  & 99.36  & \bf{99.51}  & 99.32 &  99.35 \\
							\hline
						\end{tabular}
					}
				}
			\end{center}
		\end{table}

		(3) \textit{Network Depth}: We also conduct experiments to test the impact of different network depths by reducing or adding non-linear layers. We train and test on 5 networks with depths 5, 7, 9, 11 and 13. The experimental results are summarized in Table \ref{networkdepth}. From this table it can be seen that increasing the network depth does not always generate better results, which may be a result of the gradient vanishing problem~\cite{hochreiter1998vanishing} encountered by deeper neural networks. For this experiment, the best performance can be achieved by setting the network depth to 7, which we use as the default setting for the network depth in our experiments.
		
		\begin{table}[htp]
			\caption{\label{networkdepth} {{Overall accuracy (\%) with different network depths.}}}
			\begin{center}
				{\normalsize
					\scalebox{0.85}[0.85]
					{
						\begin{tabular}{|c|c|c|c|c|c|}
							\hline
							Network Depth   & 5      & 7   & 9     & 11    & 13 \\
							\hline
							OA              & 99.35  & \bf{99.52} & 99.48 & 99.45  & 99.41 \\
							\hline
						\end{tabular}
					}
				}
			\end{center}
		\end{table}
		
		(4) \textit{Patch Size}: We also investigate the performances of our proposed method with respect to different data patch sizes $k=\{1,3,5,9,13\}$. These results are shown in Table \ref{patchsize}. As can be seen, a larger patch size of the sample generates better results. This is because more spatial information is incorporated into the training process. However, it also takes more computation time to train the network with an increasing patch size. Thus we choose 9 as the default setting of patch size as a tradeoff between performance and the running time.
		
		\begin{table}[htp]
			\caption{\label{patchsize} {{Overall accuracy (\%) with different patch sizes.}}}
			\begin{center}
				{\normalsize
					\scalebox{0.85}[0.85]
					{
						\begin{tabular}{|c|c|c|c|c|c|}
							\hline
							Patch Size   & 1      & 3      & 5     & 9     & 13 \\
							\hline
							OA           & 95.12  & 96.40  & 98.13 & 99.43 & \bf{99.52} \\
							\hline
						\end{tabular}
					}
				}
			\end{center}
		\end{table}
		
		(5) \textit{Smoothness parameter}: Finally, we test the impact of different smoothness parameter $\mu$ on the obtained {{classification}} results. Table \ref{smoothparameter} shows these results, where it can be observed that setting $\mu=20$ produces the best {{classification}} results. Thus we use this value in our experiments.
		
		\begin{table}[htp]
			\caption{\label{smoothparameter} {{Overall accuracy (\%) with different smooth parameters.}}}
			\begin{center}
				{\normalsize
					\scalebox{0.75}[0.75]
					{
						\begin{tabular}{|c|c|c|c|c|c|c|c|}
							\hline
							Smooth Parameter& 1    & 3    & 5   & 10  & 20  & 30   & 50 \\
							\hline
							OA      & 98.43  & 98.61  & 98.70 & 98.89 & \bf{99.53} & 99.31 & 99.35\\
							\hline
						\end{tabular}
					}
				}
			\end{center}
		\end{table}
		
		For other parameters, such as the kernel size in max pooling layer which is set as 2, the number of nodes in the two fully connected layers, which are set as 200 and 100 respectively, we {{observe}} consistent performance under variations. Therefore, we {{didn't}} report their impact to the performance in this section. {{The learning rate $\alpha$ is set as 0.001 and the batch size $n_{batch}$ is set as 100 for synthetic and Indian Pines dataset since consistent performance is also observed under variations. While for Pavia University dataset the learning rate is set to 0.0001 and the batch size $n_{batch}$ is set as 150. For the following experiments, we adopt the same network structure.}}
		
		\subsubsection{Experimental result of the synthetic HSI data}
\begin{figure*}
			\centering
			\includegraphics[width=0.9\linewidth]{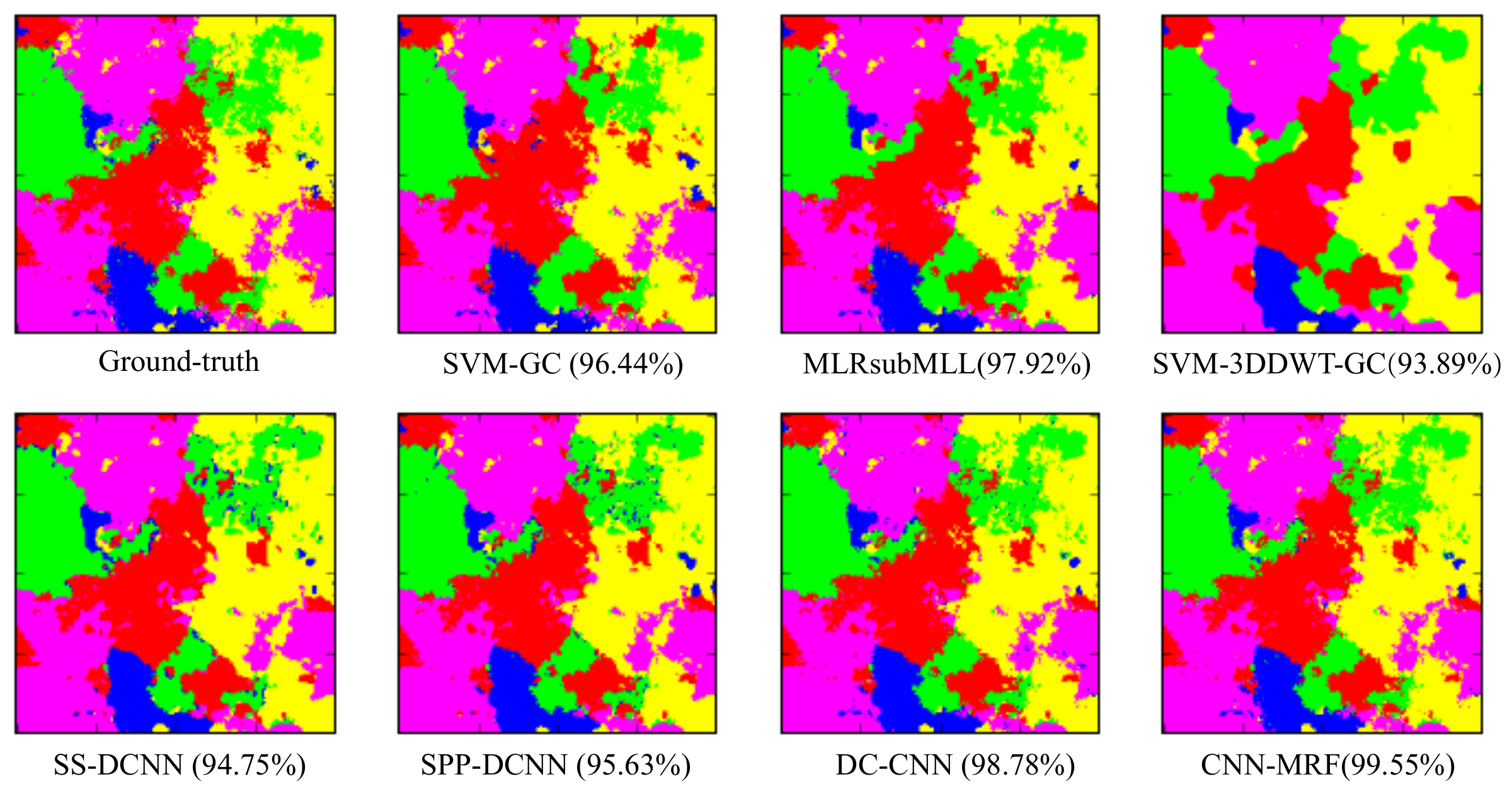}
			\caption{{{Classification maps obtained by all competing methods on the synthetic dataset (overall accuracies are reported in parentheses).}}}\label{fig_simu}
		\end{figure*}

		{{Using the parameter settings above for the CNN structure, we compare our CNN-MRF method with other competing methods. The final classification results are shown in Table \ref{table_simu}}}.  From Table \ref{table_simu}, we see that the proposed CNN-MRF method achieves better performance with respect to OA, AA and $\kappa$ than other methods. We emphasize that all non-deep learning methods {{(namely SVM-GC, MLRsubMLL and SVM-3DDWT-GC)}} in this experiment use the MRF to post-process the classification map, and the differences between them and our method are the way to extract features {{and the iteratively updating the CNN parameters and class labels.}} Thus the improvement in our method here is due to the use of CNN {{and the integration of CNN with MRF. Comparing with other deep-learning method, our CNN-MRF method also outperforms them. For better visualization, we also demonstrate the final classification map in Figure \ref{fig_simu}.}} From Figure \ref{fig_simu}, we can see that the classification 
maps obtained by our method are visually closer to the ground truth map than the other methods, which is consistent with the quantitative metric OA. 
		
		\begin{table*}[htp]
			\caption{\label{table_simu} {{Overall accuracies (\%), average accuracies (\%), kappa statistics and running time of all competing methods on the synthetic dataset.}}}
			\begin{center}
				{\normalsize
					\scalebox{0.75}[0.75]
					{
						\begin{tabular}{|c||c|c|c|c|c|c|c|}
							\hline
							Class   & SVM-GC~\cite{tarabalka2010svm}   & MLRsubMLL~\cite{li2012spectral}  & SVM-3DDWT-GC~\cite{cao2017integration}  &  SS-DCNN~\cite{yue2015spectral} & SPP-DCNN~\cite{yue2016deep} & DC-CNN~\cite{zhang2017spectral} & CNN-MRF \\ 
							\hline
							OA      & 96.44    & 97.92      & 93.89     &    94.75  &  95.63    & 98.78 & \bf{99.55}  \\ 
							\hline
							AA      & 94.55    & 95.74      & 91.37     &    92.46  &  93.24   & 96.59  & \bf{97.83}  \\ 
							\hline
							$\kappa$   & 95.23    & 96.03      & 92.84     & 93.57   & 94.46   & 97.21  & \bf{98.26}  \\ 
							\hline 
						\end{tabular}
					}
				}
			\end{center}
		\end{table*}	
		
		\begin{table}[htp]
			\caption{\label{table1} Statistics of the Indian Pines data set, including the name, the number of training, test and total samples for each class.}
			\begin{center}
				{\normalsize
					\scalebox{0.87}[0.87]
					{
						\begin{tabular}{|c|c|c|c|c|}
							\hline
							\multicolumn{2}{|c}{Class} & \multicolumn{3}{|c|}{Samples}\\
							\hline
							No & Name & Train & Test & Total\\
							\hline
							1 & Alfalfa & 5   & 41   & 46   \\
							2 & Corn-no till & 143 & 1285 & 1428 \\
							3 & Corn-min till & 83  & 747  & 830  \\
							4 &  Corn   & 24  & 213  & 237  \\
							5 & Grass-pasture & 49  & 434  & 483  \\
							6 & Grass-trees   & 73  & 657  & 730  \\
							7 & Grass-pasture-mowed & 3 & 25 & 28 \\
							8 & Hay-windrowed   & 48  & 430  & 478  \\
							9 &  Oat    & 2  & 18    & 20\\
							10 & Soybean-no till & 98 & 847 & 972\\
							11 & Soybean-min till &	246 & 2209 & 2455\\
							12 & Soybean-clean  & 60 & 533 & 593\\
							13 & Wheat & 21 & 184 & 205\\
							14 & Woods & 127 & 1138 & 1265\\
							15 & Buildings-Grass-Trees-Drives & 39 & 347 & 386\\
							16 & Stone-Steel-Towers	& 10 & 83 & 93\\
							\hline
							\multicolumn{2}{|c|}{Total} & 1025 & 9224 & 10249\\
							\hline
						\end{tabular}
					}
				}
			\end{center}
		\end{table}

		\subsection{Real HSI Data}
		We next test our method on two real HSI benchmark datasets, the Indian Pines dataset and Pavia University dataset, in a variety of experimental settings.
		
		\begin{table*}[t!]
			\caption{\label{table2} {{Individual class, overall, average  accuracies (\%) and kappa statistics of all methods on the Indian Pines image test set.}}}
			\begin{center}
				{\normalsize
					\scalebox{0.85}[0.85]
					{
						\begin{tabular}{|c||c|c|c|c|c|c|c|c|}
							\hline
							\multirow{2}{*}{Class} &
							\multicolumn{4}{c|}{Classification algorithms} &
							\multicolumn{4}{c|}{{{Regularized Classification}} algorithms} \\
							\cline{2-9}
							& SVM       & MLRsub & {{SVM-3DDWT}} & CNN & SVM-GC~\cite{tarabalka2010svm} & MLRsubMLL~\cite{li2012spectral} & {{SVM-3DDWT-GC}}~\cite{cao2017integration} &  CNN-MRF \\ 
							\hline
							1	& 73.17 & 46.34 & 63.41  & 85.23 & \bf{95.12}  & \bf{95.12} & 82.93  & 86.52 \\ 
							2	& 62.65 & 40.93 & 89.81  & 90.17 & 68.48  & 50.04  & \bf{95.56} & 91.46 \\ 
							3	& 52.88 & 26.24 & 91.97  & 93.43 & 56.49  & 13.12   & 95.72 &  \bf{96.35} \\ 
							4	& 32.39 & 17.37 & 80.75  & 84.19 & 77.00  & 15.02  & 95.77 & \bf{96.22} \\ 
							5	& 91.24 & 70.97 & 96.77  & 98.76 & 94.47  & 73.04   & 96.54 &  \bf{99.48} \\ 
							6	& 92.09 & 94.37 & 98.78 &  99.43 & 97.72  & 98.93   & 99.39 &  \bf{99.82} \\ 
							7	& 36.00 & 18.18 & 56.00 &  74.52 & 34.42   & 37.25  & 0  &  \bf{78.00} \\ 
							8	& 95.58 & 96.51 & \bf{100} & 98.74 & \bf{100}  & \bf{100}    & \bf{100}  & 98.84 \\ 
							9	& 0     & 22.22 & 94.44  &  \bf{100}   & 0    & 0      & 88.89    & \bf{100} \\ 
							10	& 61.44 & 25.06 & 88.44 &  91.71 & 75.06 & 19.68  & 92.22 &  \bf{94.26} \\ 
							11	& 86.92 & 78.23 & 95.79 &  95.59 & 95.47 & 88.82  & \bf{98.05} &  96.48 \\ 
							12	& 76.36 & 16.51 & 94.93 &  89.97 & \bf{99.44} & 16.51  & 98.31 &  91.85 \\ 
							13	& 91.85 & 93.48 & 94.57 &  98.64 & 98.37 & \bf{99.46}    & 98.37 &  98.85 \\ 
							14	& 97.01 & 99.38 & 97.72 &  97.88 & 97.45 & \bf{99.91}    & 99.03 &  98.36 \\ 
							15	& 48.13 & 4.32  & 79.83 &  89.95 & 76.66 & 60.52  & 89.91 &  \bf{91.54} \\ 
							16	& 91.57 & 77.11 & 75.90 &  96.83 & \bf{98.80} & 83.13  & 72.29 &  97.85 \\ 
							\hline
							OA  & 77.02 & 63.12 & 93.19 &  94.05 & 85.92 & 70.45  & 94.28 &  \bf{96.12}  \\ 
							AA	& 68.08 & 50.57 & 87.44 &  92.97 & 76.91 & 56.82  & 87.69 &  \bf{94.75} \\ 
							Kappa	& 73.49 & 53.13 & 92.22 &  92.86 & 83.78 & 65.43  & 94.85 &  \bf{95.78} \\ 
							\hline
						\end{tabular}
					}
				}
			\end{center}
		\end{table*}	
		
		\subsubsection{AVIRIS Indian Pines Data}
\begin{figure}
			\centering
			\includegraphics[width=0.75\linewidth]{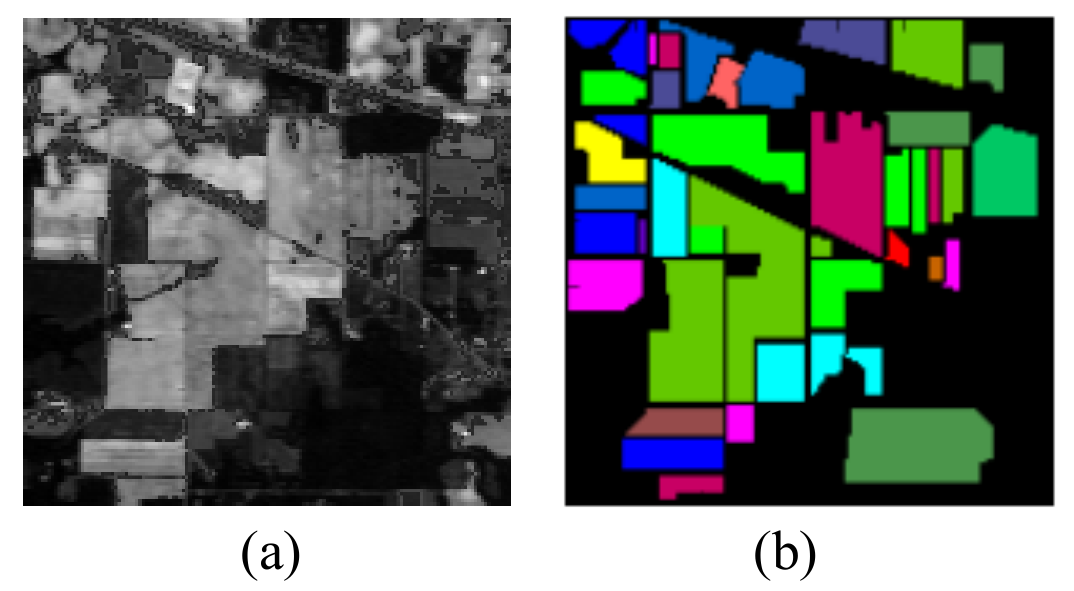}
			\caption{{{Indian Pines image and related ground truth categorization information. (a) The original HSI. (b) The ground truth categorization map (This figure is better seen by zooming on a computer screen.)}}}\label{indianpines_Image}
		\end{figure}

		This data set was gathered by the Airborne Visible/Infrared Imaging Spectrometer (AVIRIS) sensor over the Indian Pines test site in North-western Indiana in June 1992. The original dataset contains 220 spectral reflectance bands in the wavelength range 0.4$-$2.5 {{$\mu$m}}, of which 20 bands cover the region of water absorption. Unlike other methods which remove the 20 polluted bands, we keep all 220 bands in our experiments. This HSI has a spectral resolution of 10 {{nm}} and a spatial resolution of 20 {{m}} by pixel, and the spatial dimension is 145$\times$145. The ground truth contains 16 land cover classes. This dataset poses a challenging problem because of the significant presence of mixed pixels in all available classes and also because of the unbalanced number of available labeled pixels per class. A sample band of this dataset and the related ground truth categorization map are shown in Figure \ref{indianpines_Image}. 
		
		To first evaluate our proposed CNN-MRF method in the scenario of limited training samples, we randomly choose $10\%$ of the available labeled samples for each class from the reference data, which is an imbalanced training sample case, and use the remaining samples in each class for testing. The training and testing sets are summarized in Table \ref{table1}. This experiment is repeated 20 times for each method and the average performance is reported. 
		
		Additionally, some parameters need to be set in advance in these experiments. For the competing methods, their parameters are set as their papers suggest: For SVM-based methods, the RBF kernel parameter $\gamma$ and the penalty parameter $C$ are tuned through 5-fold cross validation ($\gamma=2^{-8}, 2^{-7},\dots,2^{8}$, $C=2^{-8}, 2^{-7},\dots,2^{8}$). For SVM-GC and SVM-3DDWT-GC, the spatial smoothness parameter $\beta$ is set as 0.75 advised by~\cite{cao2017integration,tarabalka2014graph}. For MLRsubMLL, the smoothness parameter $\mu$ and the threshold parameter $\tau$ are set following~\cite{li2012spectral}. For our proposed CNN-MRF method, {{we adopt the same network structure as the synthetic HSI data.}}
		
\begin{figure*}
			\centering
			\includegraphics[width=1\linewidth]{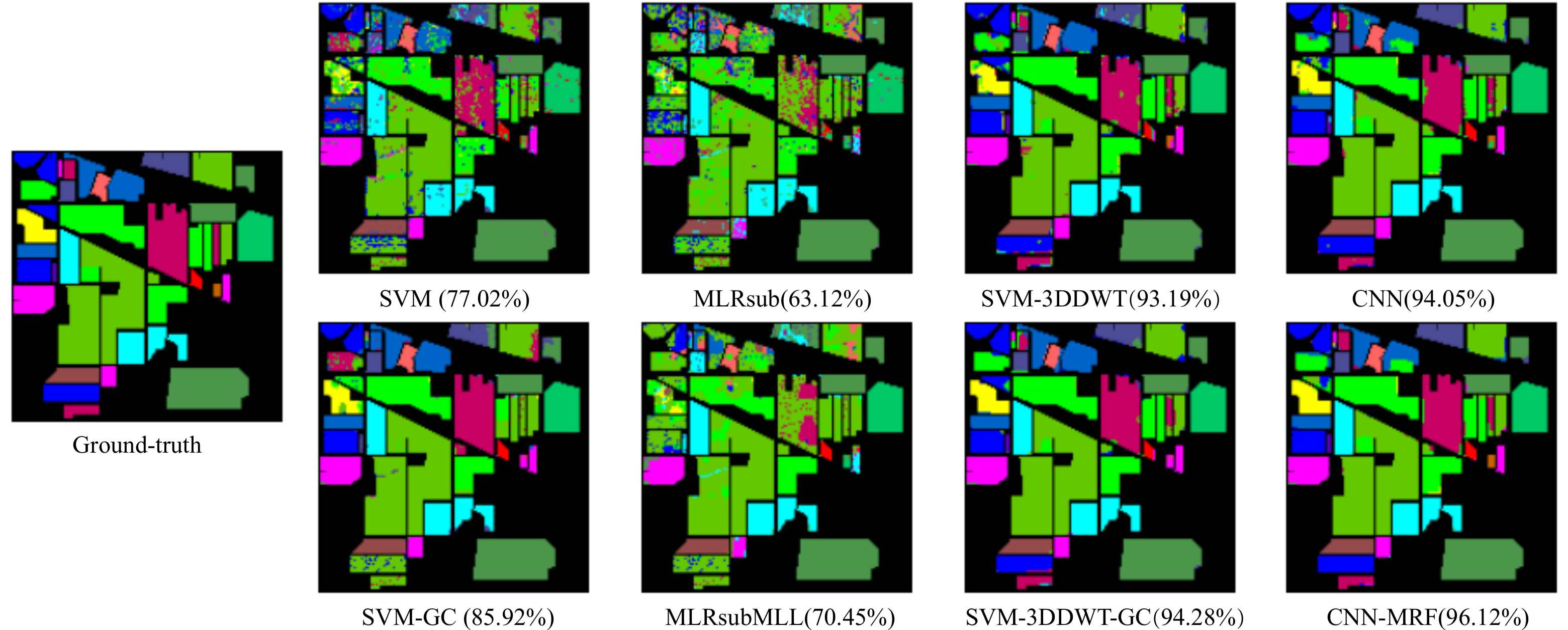}
			\caption{{{Classification maps obtained by all methods on the Indian Pines dataset (overall accuracies are reported in parentheses).}}}\label{fig4}
		\end{figure*}

		For the above experimental settings, in order to measure the performance improvement due to including spatial contextual information with the MRF, we also report the classification results of each method \textit{without} using the MRF prior, and call these methods SVM, MLRsub, {{SVM-3DDWT}} and CNN, respectively. Classification maps on the Indian dataset are shown for all methods in Figure \ref{fig4}, and the accuracies (i.e. individual class accuracy, OA, AA and $\kappa$) are reported in Table \ref{table2}.

		From Table \ref{table2}, we highlight two main results. First, CNN-MRF and CNN, which are both deep learning methods, achieve the first and third best performance in terms of the three criteria (OA, AA and $\kappa$); CNN-MRF attains about 2\% improvement in term of OA in this scenario, with {{SVM-3DDWT-GC}} trailing slightly behind (94.28\%). {{And CNN falls behind SVM-3DDWT-GC only 0.23\%. This means that using a CNN for classification without the MRF performs better than a MRF-based model with other classifier except the SVM classifier with 3DDWT features.}} The CNN therefore significantly helps for this problem. Moreover, as depicted in Figure  \ref{fig4}, the classification maps of the CNN-based methods are noticeably closer to the ground truth map. Finally, directly comparing the MRF and non-MRF based methods, we can conclude that using an MRF prior significantly improves the classification accuracy of any particular classifier because it further embeds the spatial smoothness 
		information into the segmentation stage. Therefore, the superior performance of our proposed CNN-MRF method can be explained by using the CNN and MRF strategies simultaneously to fully exploit the spectral and spatial information in a HSI.
		
		
		
		\subsubsection{ROSIS Pavia University Data}
\begin{figure}
			\centering
			\includegraphics[width=0.75\linewidth]{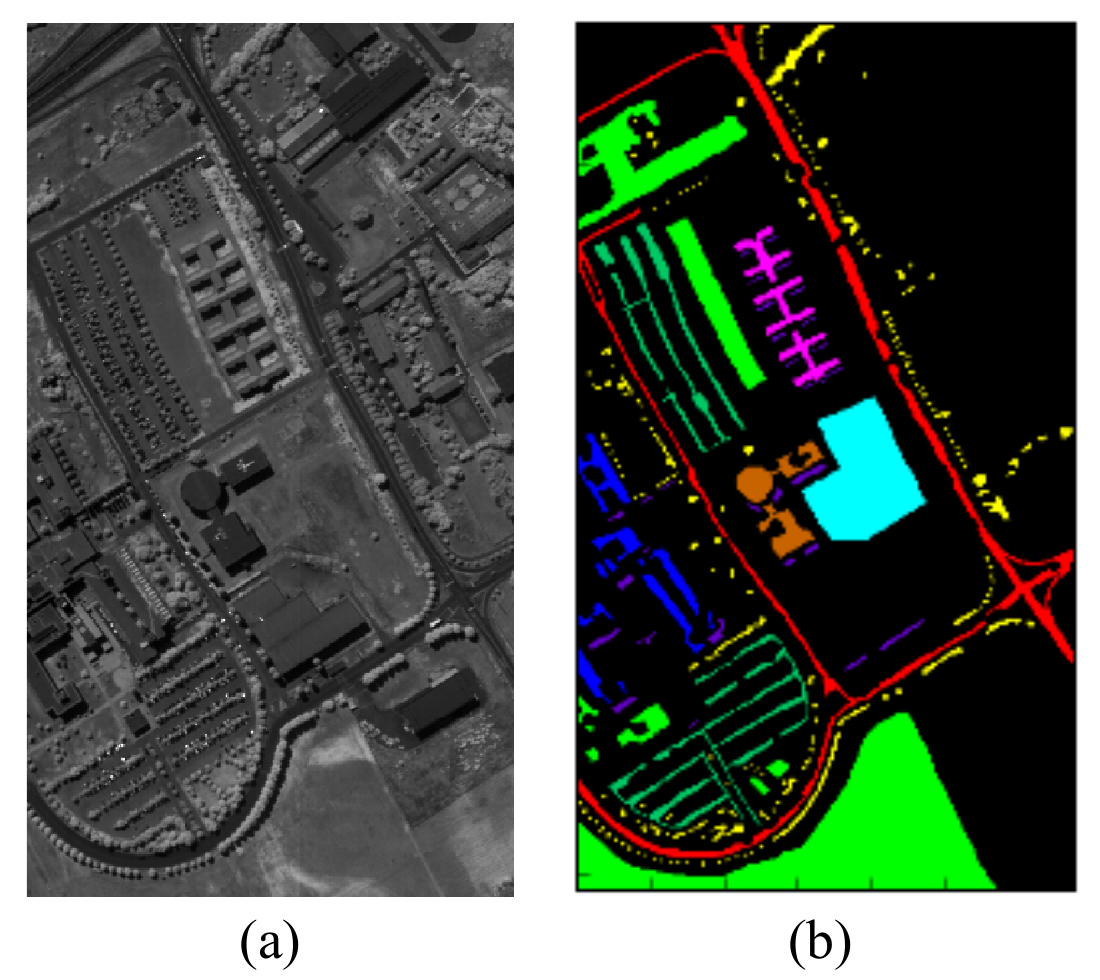}
			\caption{{{Pavia University image and related ground truth categorization information. (a) The original HSI. (b) The ground truth categorization map. (This figure is better seen by zooming on a computer screen.)}}}\label{fig6}
		\end{figure}

		We perform similar experiments on a second real HSI dataset. This HSI was acquired by the Reflective Optics System Imaging Spectrometer (ROSIS) over the urban area of the University of Pavia in northern Italy on July 8, 2002. The original dataset consists of 115 spectral bands ranging from 0.43 to 0.86 {{$\mu$m}}, of which 12 noisy bands are removed and only 103 bands are retained in our experiments. The scene has a spatial resolution of 1.3 {{m}} per pixel, and the spatial dimension is 610$\times$340. There are 9 land cover classes in this scene and the number of each class is displayed in Table \ref{table3}. We show a sample band and the corresponding ground truth class map in Figure  \ref{fig6}.

	\begin{table}[htp]
		\caption{\label{table3} Statistics of the Pavia University data set, including the name, the number of training, test and total samples for each class.}
		\begin{center}
			{\normalsize
			\scalebox{0.85}[0.85]
				{
					\begin{tabular}{|c|c|c|c|c|}
						\hline
						\multicolumn{2}{|c}{Class} & \multicolumn{3}{|c|}{Samples}\\
						\hline
						No & Name & Train & Test & Total\\
						\hline
						1 & Asphalt &	40 & 6591  & 6631  \\
						2 & Meadows	&  40 & 18609  & 18649\\
						3 & Gravel	& 40 & 2059  & 2099\\
						4 & Trees	& 40 & 3024  & 3064\\
						5 & Painted metal sheets &	40 & 1305 & 1345\\
						6 & Bare Soil &	40 & 4989 & 5029\\
						7 & Bitumen	& 40 & 1290 & 1330\\
						8 & Self-Blocking Bricks &	40 & 3642 & 3682 \\
						9 & Shadows & 40 & 907 & 947\\
						\hline
						\multicolumn{2}{|c|}{Total} & 360 & 42416 & 42776\\
						\hline
					\end{tabular}
				}
			}
		\end{center}
	\end{table}		
		
		\begin{table*}[th]
			\caption{\label{table4} {{Individual class, overall, average  accuracies (\%) and kappa statistics of all competing methods on the Pavia University image test set.}}}
			\begin{center}
				{\normalsize
					\scalebox{0.85}[0.85]
					{
						\begin{tabular}{|c||c|c|c|c||c|c|c|c|}
							\hline
							\multirow{2}{*}{Class} &
							\multicolumn{4}{c|}{Classification algorithms} &
							\multicolumn{4}{c|}{{{Regularized classification}} algorithms} \\
							\cline{2-9}
							& SVM      & MLRsub& {{SVM-3DDWT}} & CNN       & SVM-GC~\cite{tarabalka2010svm}   & MLRsubMLL~\cite{li2012spectral} & {{SVM-3DDWT-GC}}~\cite{cao2017integration} & CNN-MRF \\ 
							\hline
							1& 72.17    & 62.10 & 84.78  & 96.82     & 97.74    & 87.07     & 92.03  & \bf{98.02} \\ 
							2& 64.85    & 46.57 & 93.62  & 96.80     & 67.92    & 96.97     & 97.33  & \bf{97.78} \\ 
							3& 75.57    & 58.38 & 85.04  & 86.16     & 90.38    & 77.27     & 86.89  & \bf{88.47} \\ 
							4& 89.62    & 89.75 & 95.90  & 98.54     & 90.34    & 83.90     & 97.09  & \bf{99.17} \\ 
							5& 97.78    & 99.54 & 98.77  & 99.88     & 99.85    & 99.54     & 98.70  & \bf{99.90} \\ 
							6& 72.42    & 73.80 & 94.29  & 90.40     & 94.75    &\bf{99.40} & 98.12  & 93.00 \\ 
							7& 86.82    & 80.78 & 96.59  & 86.92     & 71.16    &\bf{94.50} & 98.84  & 87.47 \\ 
							8& 67.13    & 66.56 & 79.82  & 90.94     & 68.51    & 64.83     & 84.46  & \bf{91.66} \\ 
							9& 97.57    & 99.56 &\bf{100}& 97.53     & 99.67    & 99.78     &\bf{100}& 98.03 \\ 
							\hline
							OA& 73.41   & 61.36 & 91.27  & 94.82     & 80.21    & 91.13     & 94.12  &\bf{96.18}  \\ 
							AA& 79.36   & 75.23 & 92.09  & 93.77     & 86.70    & 89.25     & 94.83  &\bf{94.83} \\ 
							Kappa& 66.23   & 53.26 & 88.55  & 93.89     & 75.09    & 88.19     & 93.55  &\bf{94.62} \\ 
							\hline
						\end{tabular}
					}
				}
			\end{center}
		\end{table*}

\begin{figure*}
			\centering
			\includegraphics[width=0.85\linewidth]{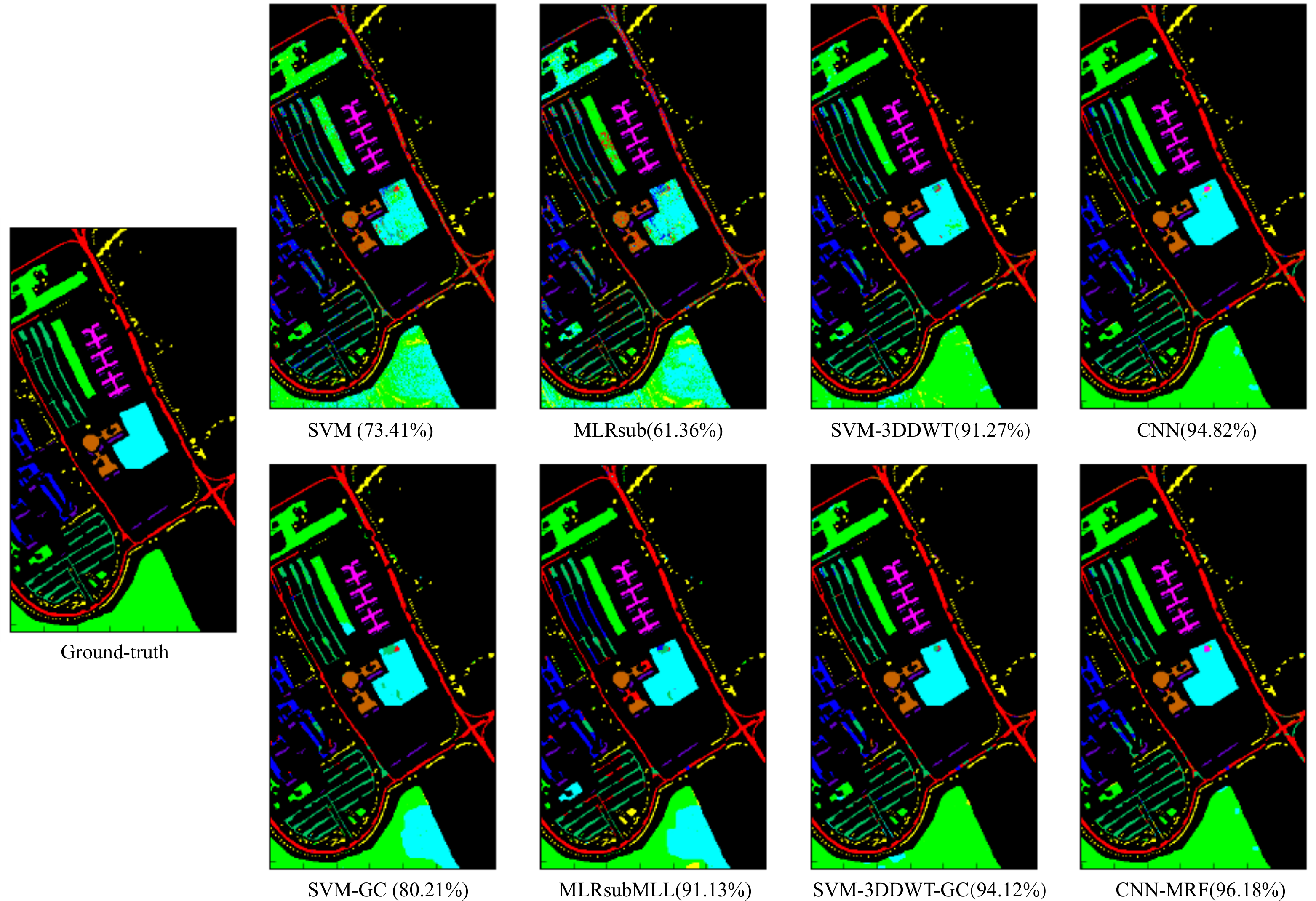}
			\caption{{{Classification maps obtained by all competing methods on the Pavia University dataset (overall accuracies are reported in parentheses).}}}\label{fig7}
		\end{figure*}

		To evaluate the performance of our proposed CNN-MRF method using only a small number of labeled training samples, we randomly chose 40 samples for each class from the ground truth data for training, which gives a balanced training sample, and the remaining samples are used for testing. The related statistics are also summarized in Table \ref{table3}. As previously, we repeat this experiment 20 times for each method and report the average performance. All parameters involved in the compared methods are tuned in the same way as the previous Indian Pines experiment. The network structure settings of the CNN are also the same. Classification and segmentation maps obtained by all methods on this dataset are illustrated in Figure  \ref{fig7} and accuracies (i.e. individual class accuracy, OA, AA and $\kappa$) are summarized in Table \ref{table4}.

		\begin{figure*}[t]
			\centering
			\includegraphics[width=0.8\linewidth]{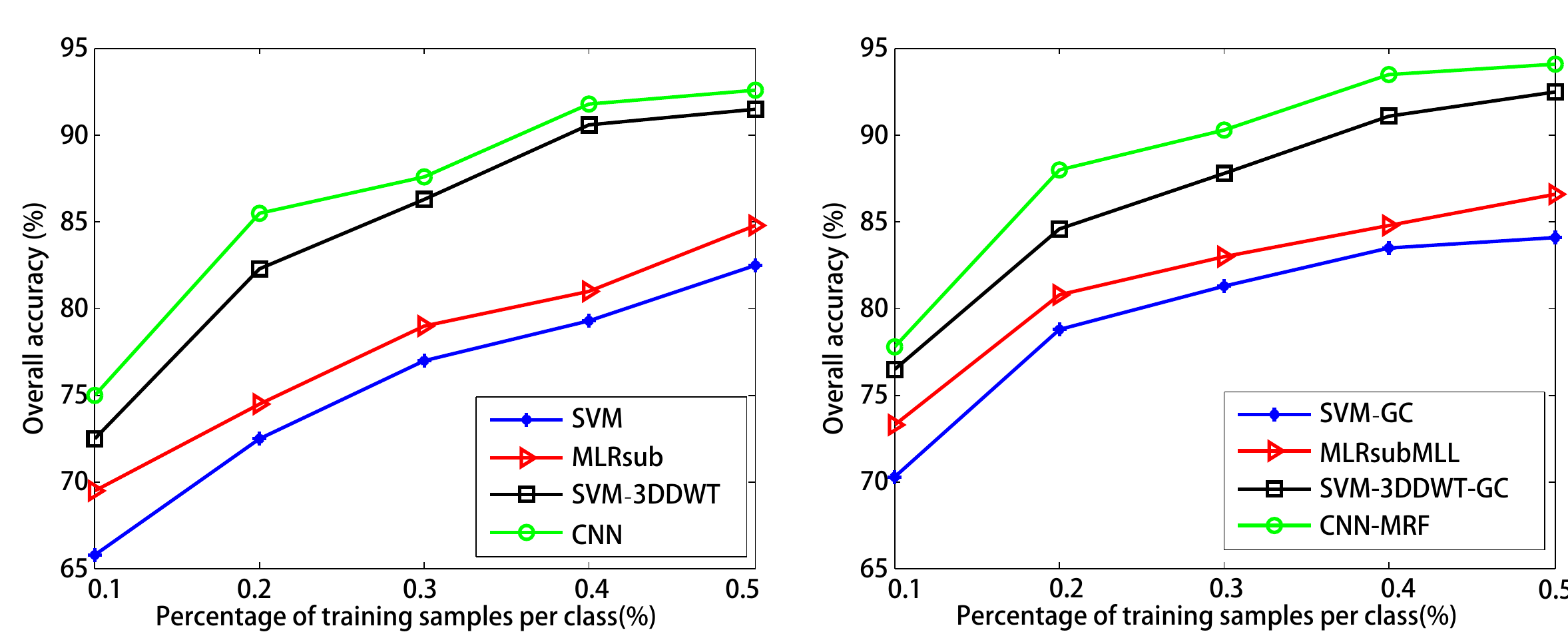}
			\caption{{{Overall accuracy (\%) obtained by all competing methods with different proportions of training samples on Pavia University dataset. (a) Classification results. (b) {{Regularized classification}} results.}}}\label{fig8}
		\end{figure*}
		
		From Table \ref{table4}, we can conclude that, for this dataset, our CNN-MRF approach again achieves the best performance in terms of the three quantitative criteria. It is also worth noting that the CNN without MRF again performs third best with respect to OA and $\kappa$, which means that the CNN plays {{an important}} role in improving the classification accuracy. By using the MRF prior, CNN-MRF obtains about 2\% improvement in terms of the OA compared with the CNN. However, for other {{classification}} algorithms we observe that the MRF can greatly boost classification accuracy. For example, MLRsubMLL has about 30\% improvement according to OA compared with MLRsub. SVM-3DDWT-GC has the second best {{classification}} OA (94.12\%) due to the 3D discrete wavelet transform (3DDWT), as has been previously studied in~\cite{cao2017integration}. Meanwhile, it can be seen from Figure  \ref{fig7} that CNN-MRF obtains much smoother {{classification}} map than other 
		methods, which is consistent with the results shown in Table \ref{table4}. Consequently, improvement of our CNN-MRF approach can be explained by the use of CNN and MRF models simultaneously. 
		
		
		\subsection{Limited training data scenarios}
		To analyze the sensitivity of the proposed method to training sets consisting of limited training samples, we conduct additional experiments in which 0.1\%, 0.2\%, 0.3\%, 0.4\% and 0.5\% of each class are randomly selected from the Pavia University data as training samples and the remaining are used for testing. For this experiment, we adopt the data augmentation technique~\cite{krizhevsky2012imagenet} to help the training process of the CNN. The OA of all methods are displayed in Figure  \ref{fig8}. From Figure \ref{fig8}(a), we observe that the classification results of the CNN method outperform the other methods for each training set size. Additionally, when the spatial prior is considered using the MRF, the {{classification}} results in Figure \ref{fig8}(b) significantly improve the corresponding classification results in Figure  \ref{fig8}(a), again indicating that the MRF is an important factor for improving the classification accuracy.
		
		\subsection{Study on the interaction between CNN and MRF}
		{{Furthermore, in order to clearly illustrate the interactive influence between CNN and MRF, we also report the OA of our method on the three datasets as a function of iteration. The results are shown in Table \ref{Epoch_goes}. Specifically, in our experiments, we first train our CNN for 30 epochs using the training data $\mathcal{D}_{l}$. Then, we update the class labels $\widehat{\mathbf{y}}$ every 10 epochs. From Table \ref{Epoch_goes}, it can be seen that the OA first increases quickly in the first 30 epochs. Then, the OA increases slowly for about 40 epochs. Finally, the OA decreases a little or has a slight fluctuation. Therefore, we conclude that the interaction between the CNN and MRF can help the final classification OA. In our experiments, we set the maximum epoch as 60 and report the output as the results of our method.}}
		
		\begin{table}[htp]
			\caption{\label{Epoch_goes} {{The change of overall accuracy (\%) as epoch goes.}}}
			\begin{center}
				{\normalsize
					\scalebox{0.60}[0.60]
					{
						\begin{tabular}{|c|c|c|c|c|c|c|c|c|c|c|}
							\hline
							Epoch & 10  & 20 & 30 & 40 & 50 & 60 & 70  & 80 & 90 & 100\\
							\hline
							Synthetic  & 92.53  & 96.72   & 98.85  & 99.44 & 99.52 & 99.58 & 99.44  & 99.30 & 99.12 & 99.25\\
							\hline
							Indian Pines  & 87.29  & 92.67 & 94.05 & 95.36 & 95.91 & 96.12 & 95.93 & 96.03 & 95.98 & 95.97\\
							\hline
							PaviaU  & 83.71  & 92.71 & 94.82 & 95.71 & 95.92 & 96.18 & 96.17 & 96.18 & 96.16 & 96.18\\
							\hline
						\end{tabular}
					}
				}
			\end{center}
		\end{table}

		\subsection{Comparison with other deep learning methods }\label{sec.otherdeep}
		In order to further evaluate the performance of our proposed CNN-MRF method, we compare with three recent deep learning HSI classification algorithms: SS-DCNN~\cite{yue2015spectral}, SPP-DCNN~\cite{yue2016deep} and DC-CNN~\cite{zhang2017spectral}. We use the two previous real datasets for comparison. For fair comparison, we choose the training samples with same proportion as was done in~\cite{zhang2017spectral}. Specifically, for the Indian Pines dataset, we choose 10\% samples from each class as training set, and for the Pavia University dataset we choose 5\% samples from each class as training data. We also apply the data augmentation strategy in the two experiments since all deep learning algorithms can adopt this strategy. The results for each algorithm are displayed in Table \ref{Indian-cnn-based-results} and Table \ref{paviaU-cnn-based-results}.\footnote{The results for SS-DCNN, SPP-DCNN and DC-CNN are taken from~\cite{zhang2017spectral}. The running time of Table \ref{table2} and Table {\ref{Indian-cnn-based-results}} is different since data augmentation strategy is adopted in this experiment of Table \ref{Indian-cnn-based-results}.} Results for Indian Pines are shown in Table
		\ref{Indian-cnn-based-results}, where it can be seen that our proposed CNN-MRF method achieves improved performance compared with other deep methods in terms of OA, AA and $\kappa$. We show results for the Pavia University data in Table \ref{paviaU-cnn-based-results}. {{We again observe that our method achieves the best performance compared with other methods. From Table~\ref{Indian-cnn-based-results} and ~\ref{paviaU-cnn-based-results}, we can conclude that the running time of our proposed method is a potential drawback. Here the running time reported in this experiment includes both training and testing time. (This could be further addressed with high-performance computing resources and techniques.)}}
		
		\begin{table}[htp]
			\caption{\label{Indian-cnn-based-results} {{Overall accuracy, average accuracy and kappa coefficient (\%) of all the deep learning methods on the Indian Pines dataset.}}}
			\begin{center}
				{\normalsize
					\scalebox{0.75}[0.75]
					{
						\begin{tabular}{|c|c|c|c|c|}
							\hline
							& SS-DCNN~\cite{yue2015spectral}  & SPP-DCNN~\cite{yue2016deep}     & DC-CNN~\cite{zhang2017spectral}     & CNN-MRF \\
							\hline
							OA       & 90.76    & 91.60    & 98.76      &  \bf{99.32} \\
							\hline
							AA       & 85.52    & 93.96    & 98.50      &  \bf{99.27}\\
							\hline
							kappa($\kappa$)  & 89.44    & 90.43    & 98.58  &  \bf{99.21} \\
							\hline
							Time     & 255.32   & 328.18   & 4860.57    &  1454.62\\
							\hline
							
						\end{tabular}
					}
				}
			\end{center}
		\end{table}
		
		\begin{table}[htp]
			\caption{\label{paviaU-cnn-based-results} {{Overall accuracy, average accuracy and kappa coefficient (\%) of all the deep learning methods on the Pavia University dataset.}}}
			\begin{center}
				{\normalsize
					\scalebox{0.75}[0.75]
					{
						\begin{tabular}{|c|c|c|c|c|}
							\hline
							& SS-DCNN~\cite{yue2015spectral}  & SPP-DCNN~\cite{yue2016deep}    & DC-CNN~\cite{zhang2017spectral}   & CNN-MRF \\
							\hline
							OA       & 93.34    & 94.88    & 99.68      & \bf{99.71} \\
							\hline
							AA       & 92.20    & 93.29    & 99.50      & \bf{99.55} \\
							\hline
							kappa($\kappa$)  & 91.95    & 93.21    & 99.58  & \bf{99.63} \\
							\hline
							Time     & 266.11   & 333.90   & 3421.25    & 1156.43\\
							\hline
							
						\end{tabular}
					}
				}
			\end{center}
		\end{table}

\begin{figure*}
			\centering
			\includegraphics[width=0.9\linewidth]{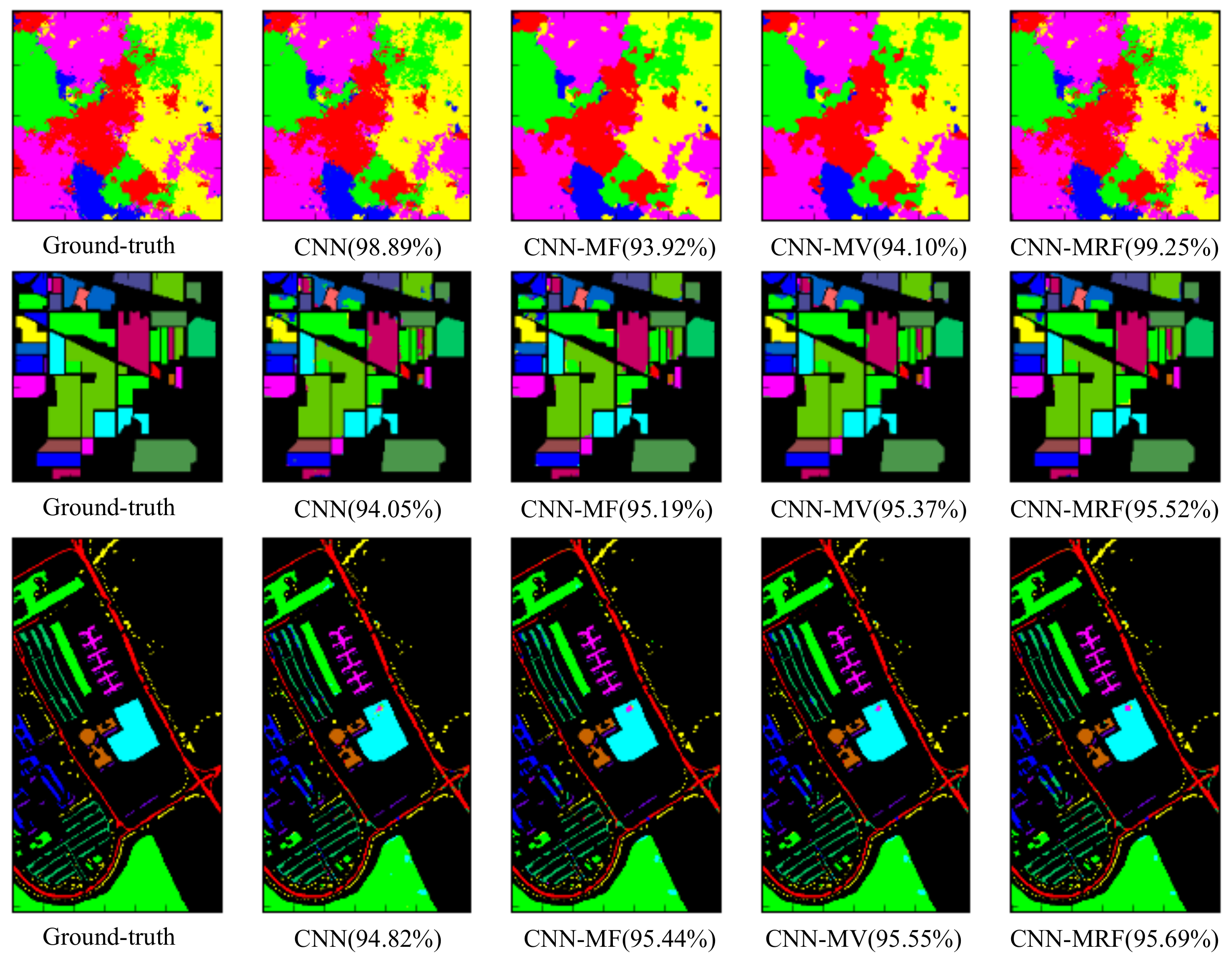}
			\caption{{{Classification maps obtained by all competing methods on the Pavia University dataset (overall accuracies are reported in parentheses).}}}\label{ppcp}
		\end{figure*}

		{{\subsection{Comparison with other label regularization methods}\label{sec.otherpostprocess}
				In order to evaluate the performance of MRF, we compare it with other label regularization methods, such as median filter~\cite{Huang1979A} and majority voting~\cite{Boyer1991MJRTY}. The other two methods are implemented just by replacing the MRF optimization with median filter method and majority voting method in our algorithm framework (we call the two methods as CNN-MF and CNN-MV respectively). All the methods are compared on the above synthetic data, Indian Pines data and Pavia University data with the same experimental settings as shown earlier. For median filter and majority voting methods, we set the window size as \{3,5,7\} and report the best result. The experimental results are shown in Table \ref{label_regularization} and Figure \ref{ppcp}. From Table \ref{label_regularization} and Figure \ref{ppcp}, we can easily observe that in MRF obtains the best performance comparing with other two simple label regularization methods on all the datasets. Thus we adopt MRF as the label regularization 
method in 
				our algorithm.}}  
		
		\begin{table}[htp]
			\caption{\label{label_regularization} {{Overall accuracy (\%) of all the label regularization methods on the synthetic, Indian pines and Pavia University datasets.}}}
			\begin{center}
				{\normalsize
					\scalebox{0.70}[0.70]
					{
						\begin{tabular}{|c|c|c|c|c|}
							\hline
							& CNN     & CNN+Median Filter & CNN+Majority Voting  & CNN-MRF \\
							\hline
							Synthetic & 98.89   & 93.92    &  94.10     & \bf{99.25} \\
							\hline
							Indian Pines  & 94.05   & 95.19    &  95.37     & \bf{95.52} \\
							\hline
							PaviaU    & 94.82   & 95.44    &  95.55     & \bf{95.69} \\
							\hline
							
						\end{tabular}
					}
				}
			\end{center}
		\end{table}
		
		\section{Conclusions}
		In this paper, we proposed a novel technique for HSI {{classification}} that incorporates both spectral and spatial information in a unified Bayesian framework. Specifically, we use a convolutional neural network (CNN) in combination with a Markov random field to classify HSI pixel vectors in a way fully takes spatial and spectral information into account. We then efficiently learn the {{classfication}} result {{by iteratively updating the CNN parameters and the class labels of all HSI pixel vectors}}. Experimental results on one synthetic HSI dataset and two real benchmark HSI datasets show that our method outperforms state-of-the-art methods, including deep and non-deep models. In the future, we will further consider the HSI classification task in unsupervised settings. Besides, we will also try to extend our model to the popular deep generative models, such as variational autoencoder (VAE)~\cite{kingma2013auto} and generative adversarial network (GAN)~\cite{
goodfellow2014generative}. {{Furthermore, we hope to design more powerful regularization regimes, extending the employed MRF one, for different application scenarios of this problem in our future research. We will also consider dimensionality reduction techniques in our methods to further make the method more efficiently computed, especially on large-scaled datasets.}}

		
		\ifCLASSOPTIONcaptionsoff
		\newpage
		\fi
		\bibliographystyle{ieee}
		\bibliography{mybibfile}

	\end{document}